\documentclass[runningheads]{llncs}
\usepackage[T1]{fontenc}
\usepackage{comment}
\usepackage{amsmath}
\usepackage{graphicx}
\begin{document}
\title{Leveraging a Statistical Shape Model for Efficient Generation of Annotated Training Data: A Case Study on Liver Landmarks Segmentation}
\titlerunning{Leveraging an SSM for Training Data Generation}
%
\author{Denis Krnjaca\inst{1}\orcidID{0009-0009-5163-3282} \and
Lorena Krames\inst{1}\orcidID{ 0009-0001-0847-1427} \and 
Werner Nahm\inst{1}\orcidID{2222--3333-4444-5555}}
\authorrunning{D. Krnjaca et al.}
%
\institute{Karlsruhe Institute of Technology, Karlsruhe, Germany
\\\email{publications@ibt.edu}\\}
%
\maketitle              
\begin{abstract}
Anatomical landmark segmentation serves as a critical initial step for robust multimodal registration during computer-assisted interventions. Current approaches predominantly rely on deep learning, which often necessitates the extensive manual generation of annotated datasets. In this paper, we present a novel strategy for creating large annotated datasets using a statistical shape model (SSM) based on a mean shape only manually labeled once. We demonstrate the method's efficacy through its application to deep-learning-based anatomical landmark segmentation, specifically targeting the detection of the anterior ridge and the falciform ligament in 3D liver shapes. A specialized deep learning network was trained with 8,800 annotated liver shapes generated by the SSM. The network's performance was evaluated on 500 unseen synthetic SSM shapes, yielding a mean Intersection over Union of 91.4\%  (87.4\% for the anterior ridge and 87.6\% for the falciform ligament). Subsequently, the network was applied to clinical patient liver shapes, with qualitative evaluation indicating promising results, highlighting the generalizability of the proposed approach. Our findings suggest that the SSM-based data generation approach alleviates the labor-intensive process of manual labeling while enabling the creation of large annotated training datasets for machine learning. Although our study focuses on liver anatomy, the proposed methodology holds potential for a broad range of applications where annotated training datasets play a pivotal role in developing accurate deep-learning models.

\end{abstract}
\section{Introduction}
A current problem in research and industry is enhancing the precision of multimodal registration during computer-assisted interventions. Approaches based on an anatomical landmark segmentation~\cite{labrunie,koo,kokko,tom} show promising improvements in 2D-3D registration results. However, achieving accurate delineation of these landmarks is still essential for enhancing the precision of the registration and the time-consuming expert manual annotation of the anatomical landmarks (e.g., the anterior ridge of the liver) on the preoperative 3D model is regardless required. Manual annotation furthermore requires special software and is dependent on the annotator. Fortunately, the growing utilization of 3D point cloud acquisition methods, such as laser scanners or LIDAR, has led to an increased focus on developing approaches for processing point clouds~\cite{Elharrouss} and hence automatizing the segmentation. A variety of deep-learning network architectures for the 3D segmentation of point clouds were introduced lately, with notable examples being the PointNet model and its enhanced version, PointNet++~\cite{pointnet,pointnet+}. One limitation of PointNet is its dependence on the equal orientation of point clouds. This requirement poses challenges when dealing with 3D models segmented from CT scans, as variations in anatomy and positions during CT acquisition may not strictly adhere to this condition. Moreover, these deep learning approaches typically require a large annotated training data set. 
\\
At the same time, statistical shape models (SSM) enable the generation of a diverse set of realistic geometries that capture the variability observed within the cohort used to construct the SSM. In the past, SSMs have been applied to a variety of applications, e.g., for the 3D reconstruction from computer tomography (CT) scans~\cite{ambellan19}. Recently, SSMs were used for the generation of training data for classification purposes~\cite{schaufel,nagel,atkins,mofrad}.
\\
Consequently we propose a method using an SSM to generate an arbitrarily large annotated training dataset of different liver shapes which is then used for anatomical landmark segmentation. The process involves a single manual labeling of desired anatomical landmarks on the mean shape extracted from the SSM. The labels are transferred to all generated shapes. Subsequently, this dataset serves as the training foundation for a deep-learning network tasked with detecting anatomical landmarks on previously unseen shapes. In the presented case study, we utilize a liver SSM to facilitate the dataset generation. Two liver landmarks, the anterior ridge and the falciform ligament, were segmented on the mean shape, and the SPRIN architecture~\cite{sprin} is chosen for the deep-learning network dedicated to 3D landmark segmentation, owing to its invariant behavior with respect to rotations. Within this work, we state two hypotheses: first, that anatomical consistency exists within shapes generated by the SSM, allowing for the transfer of anatomical landmarks from the mean shape onto the generated shapes and second, that utilizing the labeled synthetic dataset is suitable for training a deep-learning network enables robust segmentation of anatomical landmarks on unseen clinical shapes. 
\\

\section{Methodology}
\begin{figure}
\includegraphics[width=\textwidth]{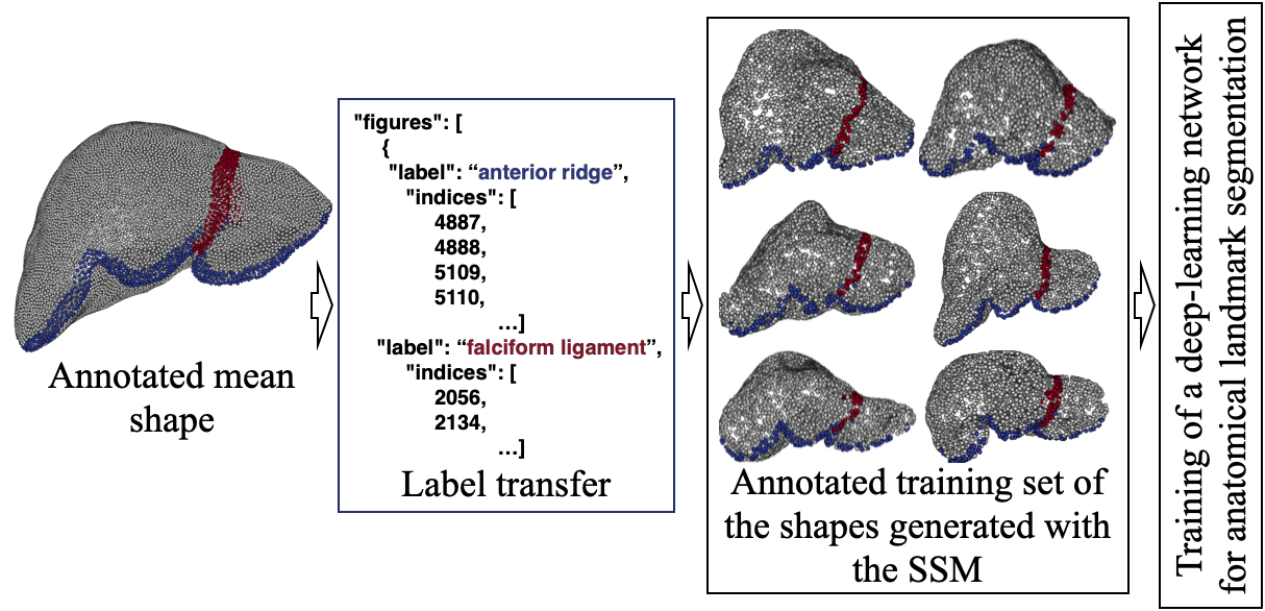}
\caption{Overview of the proposed method employing an SSM for data generation. The mean shape derived from the SSM is manually labeled once (in this example: red: anterior ridge, blue: falciform ligament). Subsequently, labels are transferred to generated shapes using vertex indices. The annotated dataset is then utilized for training a deep-learning network. } \label{overview_method}
\end{figure}
The introduced pipeline for 3D point cloud segmentation is illustrated in Fig.~\ref{overview_method}. It starts with the one-time manual annotation of the mean shape derived from the SSM. Subsequently, the labels are transferred to the training data generated from the SSM. In this process, the vertex indices, along with their corresponding labels from the mean shape, are seamlessly copied to every generated shape. This step enables the generation of an arbitrary large labeled dataset for the training of a deep-learning network for the segmentation of anatomical landmarks on 3D shapes.
\subsection{Case Study}
The presented case study serves as an illustrative application of the methodologies introduced in this work, focusing on the segmentation of two specific anatomical landmarks on 3D liver shapes. The first landmark, denoted by the blue label in Fig.~\ref{overview_method}, corresponds to the anterior ridge, while the second landmark, identified by the red label in Fig.~\ref{overview_method}, represents the falciform ligament. These landmarks were selected based on their prior use in 2D-3D registration for computer-assisted interventions in laparoscopic liver surgery~\cite{labrunie,koo}.
\\
An SSM was generated using 48 3D liver shapes, including 20 from the IRCAD dataset~\cite{ircad} according to the described method from~\cite{krnjaca}. With this SSM, it was possible to create a synthetic data set of liver shapes based on an annotated mean shape. Each generated shape is represented as
\begin{equation}
    \mathbf{x} = \mathbf{\bar{x}}+ {\Lambda}^{\frac{1}{2}}\mathbf{V} \mathbf{a} \textrm{,}
\label{gen_shapes}
\end{equation}
where $\mathbf{\bar{x}}$ is the mean shape, $\Lambda$ the eigenvalue, $\mathbf{V}$ the principal components (which is the output from the SSM generation) and $\mathbf{a}$ denotes the parameter vector allowing the linear combinations (realizations) of the principal components. When morphing the mean shape into the realizations, the annotations from the mean shape are equivalently transferred to the synthetic liver shape.
\\ 
For the automatic annotation of the anterior ridge and the falciform ligament, the \textit{SO}(3) invariant deep-learning network SPRIN~\cite{sprin} was chosen. The network takes the generated synthetic annotated point clouds from the SSM as input. Each point cloud $\mathbf{p}$ is conceptualized in SPRIN as a discrete empirical probabilistic distribution of input points as
\begin{equation}
    \mathbf{p(x)} = \frac{1}{N}\sum_{i}^{N}\delta \mathbf{(x-x_i)} \textrm{,}
\end{equation}
where $\delta$ denotes the Dirac delta function, and $\mathbf{x_i}$ represents input points. SPRIN is leveraging farthest point sampling and \textit{k}NN grouping to progressively conduct sparse rotation invariant spherical correlation between every point and its \textit{k} nearest neighbors. The structure of SPRIN is summarized in the following: The initial two sparse spherical correlation layers evaluate all input points, utilizing 2-strided 64-NN (nearest neighbors) neighborhoods. Strided indicates how often the filter matrix is moved across the input data while the correlation layers are being executed. Subsequently, the following three sparse spherical correlation layers assess 512 sampled points from the sampling \& grouping module, with 3-strided 72-NN neighborhoods, 1-strided 32-NN and 1-strided 32-NN neighborhoods, respectively.
Finally, the last three sparse spherical correlation layers evaluate 32 sampled points from the sampling \& grouping module, utilizing 1-strided 32-NN neighborhoods. The architecture design of the two feature propagation layers for segmentation remains consistent with the approach outlined in~\cite{sprin}.




\section{Experiments}
\subsection{Implementation Details}
The network SPRIN employed for the anatomical landmark segmentation is implemented in Python using the PyTorch~\cite{pytorch} framework. The Adam optimization algorithm, as described in~\cite{adam}, with a learning rate of $1e-3$ is utilized for SPRIN, employing a batch size of 12. Training was executed on a single NVIDIA RTX 2080 TI. 

\subsection{Datasets}
In total, 11,500 liver shapes were generated from the SSM by uniformly sampling $\mathbf{\alpha}$ in the range of [-2.75, 1.75] standard deviations. This range and the uniform distribution were deliberately chosen to encompass a wide variation in liver anatomy, ensuring a comprehensive representation. Following the transfer of labels from the mean shape, the shapes were down-sampled to 4,096 points, and the indices of the vertices were randomly shuffled. This step was taken to prevent the network from memorizing only the indices corresponding to specific labels. The dataset was then divided into 8,800 shapes for training, 2,200 shapes for validation, and 500 shapes for testing. 
\\
To test our hypothesis regarding the transferability of labels from the mean shape to the generated shapes from the SSM, we conducted a small annotation study. Four individuals without expertise were tasked with manually labeling two anatomical landmarks on five distinct SSM shapes. The annotations from each of the four participants were aggregated for each shape and then compared with the presumed ground truth, i.e., the transfer of labels onto the SSM shapes.  
\\
Besides the shapes generated from the SSM, liver shapes from the MedShapeNet were fed into SPRIN for a qualitative evaluation of the predictions~\cite{medshapenet}.


\subsection{Results}

\subsubsection{Evaluation of Label Transfer on the Training Data Set:} 
\begin{figure}
––\includegraphics[width=\textwidth]{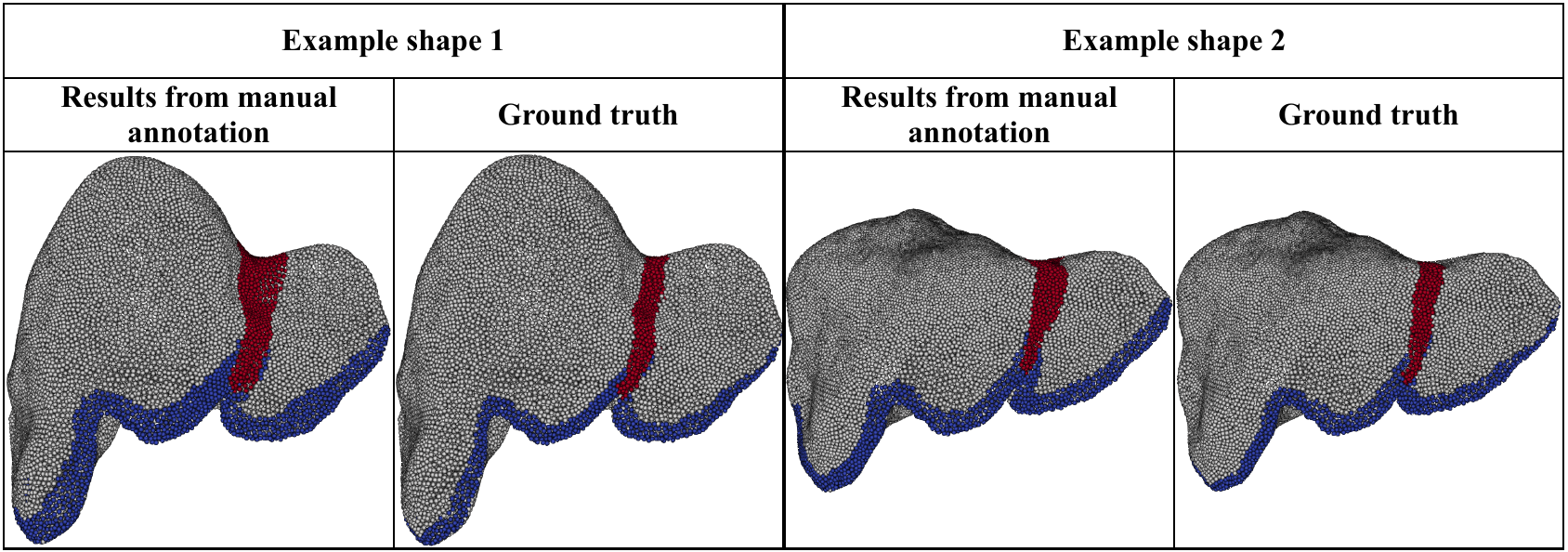 }
\caption{Two examples illustrating the aggregated labels annotated by the four individuals (first and third shape) and the corresponding results from the label transfer from the mean shape (second and last shape).}  \label{study}
\end{figure}

The evaluation employed the accuracy metric given by
\begin{equation}
\text{accuracy} = \frac{ | A \cup GT |}{ | GT |}.
\end{equation}
Here, $A$ represents the aggregated labels annotated by the four individuals, and $GT$ denotes the ground truth used. The overall accuracy was 82.45\% (averaged accuracies of all shapes), with specific accuracies of 87.05\% for the anterior ridge and 66.06\% for the falciform ligament. Two examples illustrating the study can be seen in Fig.~\ref{study}.
\\
\\
In the evaluation on the test data, the intersection over union (IoU) metric was computed by
\begin{equation}
    IoU =    \frac{ | GT \cap Pred |}{ | GT  \cup Pred |} ,
\end{equation}
where $GT$ represents the ground truth and $Pred$ denotes the prediction. This calculation was performed individually for each label within the synthetic shapes dataset. Subsequently, the IoU values were averaged for all labels to derive the mean IoU (mIoU) for each respective shape. To gain a complete overview, the mIoUs were again averaged across all shapes. The respective evaluation was performed distinguishing between the two labels. 
\begin{figure}
\includegraphics[width=\textwidth]{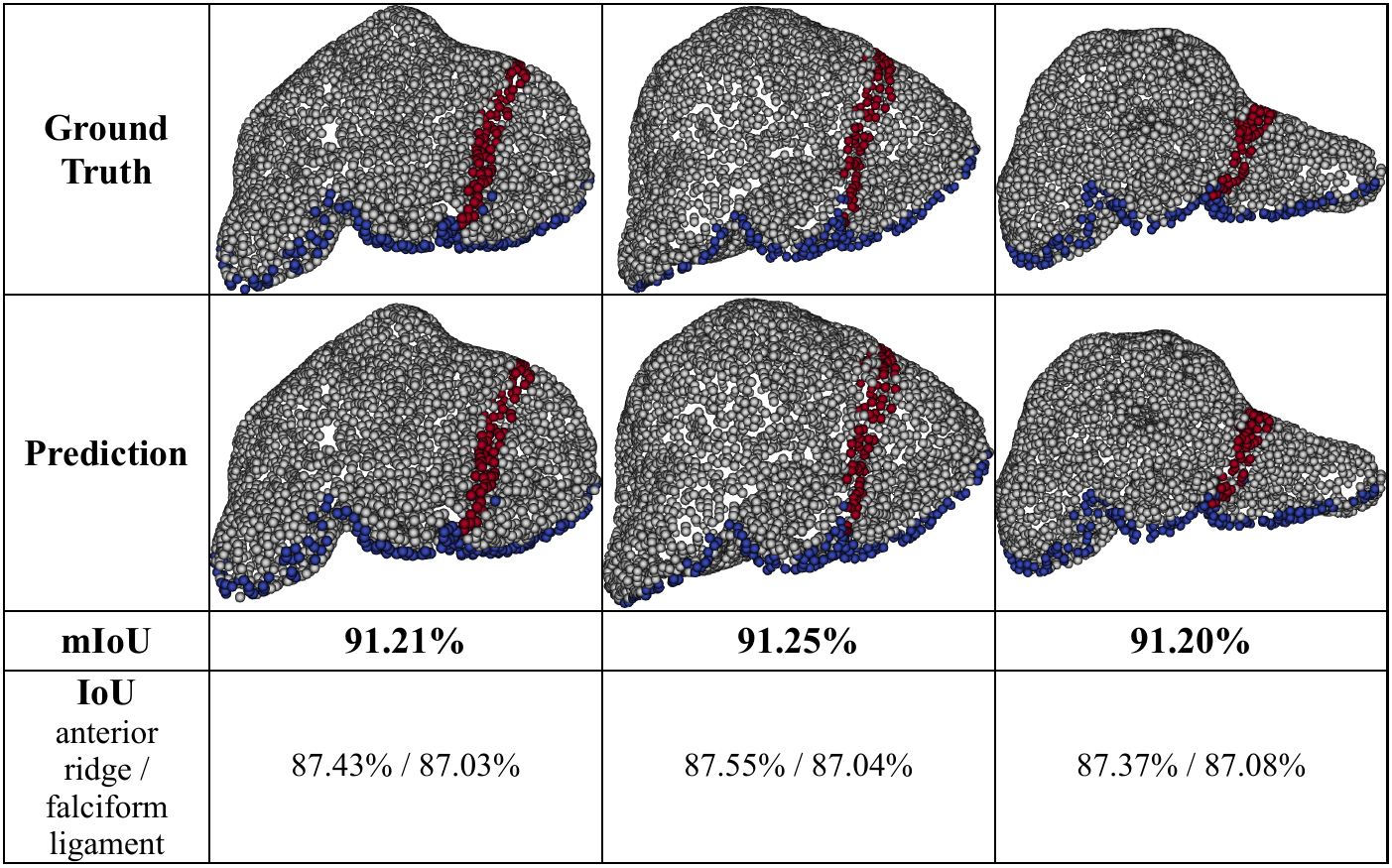 }
\caption{Evaluation of three examples from the synthetic SSM test dataset. The anterior ridge prediction is shown in blue and the falciform ligament in red. The last two rows show the mIoU over all labels as well as the IoU for the anterior ridge and the falciform ligament.}  \label{SSM_eva}
\end{figure}
\begin{figure}
\includegraphics[width=\textwidth]{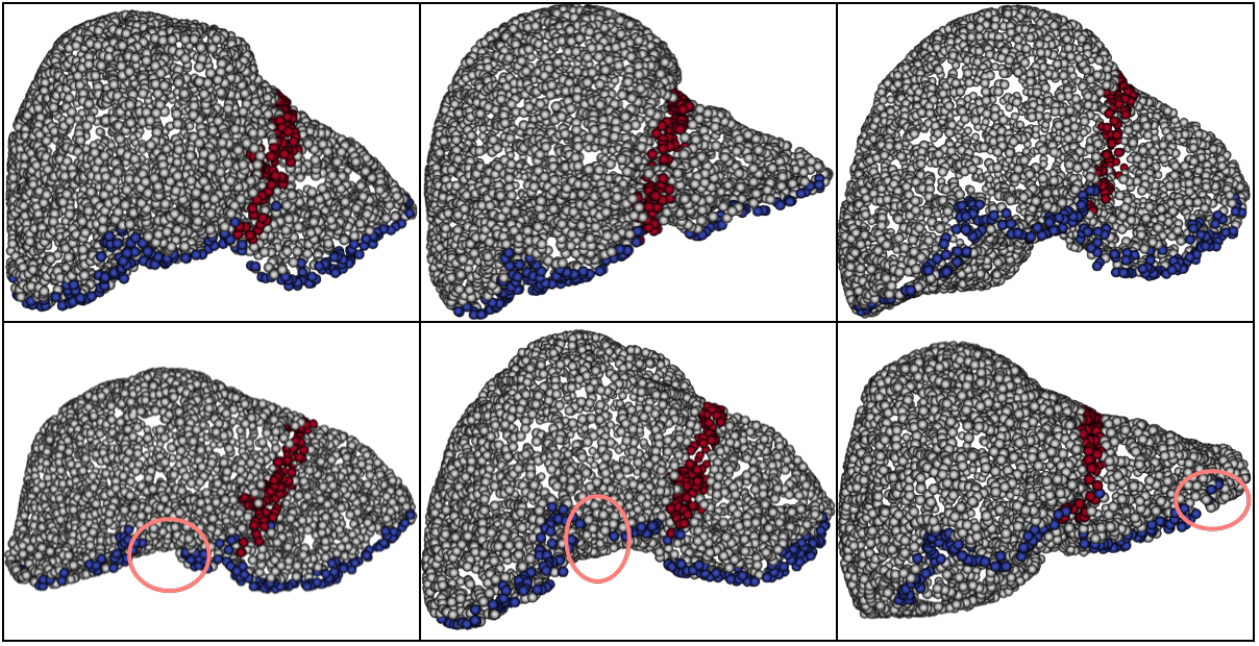 }
\caption{Qualitative evaluation on six examples from the MedShapeNet dataset~\cite{medshapenet}. The anterior ridge prediction is shown in blue and the falciform ligament in red.}  \label{MedShape_eva}
\end{figure}
\subsubsection{Quantitative and Qualitative Evaluation on SSM Shapes:}
First, the trained SPRIN model was tested on the 500 random rotated test shapes generated from the SSM. The mIoU calculated over the whole dataset was 91.4\%, with 87.4\% for
the anterior ridge and 87.6\% for the falciform ligament. Three examples can be seen in Fig.~\ref{SSM_eva}. 
\subsubsection{Qualitative Evaluation on Shapes from MedShapeNet:}
In the second experiment, the network was tested on shapes from the MedShapeNet~\cite{medshapenet}. Shapes were evaluated only if they were entirely segmented, i.e., were not showing any missing parts of the right or left liver lobe. Six examples for the prediction are shown in Fig.~\ref{MedShape_eva}. In the second row of the figure, the circles mark incomplete segmentations of the anterior ridge.

\section{Discussion and Conclusion}
The main idea we propose is leveraging an SSM to generate a comprehensive labeled training dataset for the 3D anatomical landmark segmentation by transferring point-wise annotations from the once manual labeled mean shape to the shapes generated. The underlying hypothesis is that nodes are mapped on the same anatomical structures when using the SSM transformation. This was validated by comparing the resulting labels of five shapes with the results from a four-times repated manually annotation of the same shapes. Through this investigation, evidence was furnished in support of the transferability of labels from the mean shape to the generated shapes at least for liver shapes.
\\
Additional computational resources could further improve performance. Due to the computation constraints the point clouds had to be downsampled to 4,096 points. The limited amount of points leads to a decrease of geometrical and anatomical information. Thus, using more points could increase the accuracy of the landmark segmentation.
\\
The outcomes of the case study focusing on the segmentation of two anatomical landmarks on liver shapes demonstrated high accuracy, particularly evident in the test data generated from the SSM. 
The shapes from the SSM typically exhibit an "M" shaped characteristic of the anterior ridge with two indentations (representing the round ligament and gallbladder). However, the anterior ridge does not consistently exhibit this characteristic in reality, leading to instances of incomplete segmentation, as observed in the bottom row of Fig.~\ref{MedShape_eva}. Additionally, an explicitly wide annotation occurs in areas where the gallbladder is located, as illustrated in the upper-left shape of Fig.~\ref{MedShape_eva}. 
The segmentation of the falciform ligament proved to be robust in all instances. Nevertheless, evaluating the results is not always straightforward, as the characteristic groove delineating the course of the falciform ligament is occasionally absent. For a final and reliable proof of the second hypothesis of the efficacy of labeled synthetic data for robust segmentation on clinical data more expert annotation of test shapes (e.g. from MedShapeNet)is needed. However, the starting point, represented by the round ligament, as well as the end point, corresponding to the entry and exit of the hepatic vein, were qualitatively consistent and accurately segmented. This capability suggests promising prospects for generalization across diverse patient populations, thus contributing to its possible clinical application. 
\\
An inherent challenge in anatomical landmark segmentation lies in the often indistinct borders of these landmarks. The thickness of annotations on the mean shape can significantly influence the resulting segmentation on test shapes. The increased anatomical variance across patients further compounds the difficulty of 3D landmark segmentation. Additionally, the manual or semi-automatic 3D reconstruction of point clouds from CT scans introduces variability in reconstructed shapes, sometimes leading to incomplete representations~\cite{medshapenet}. To mitigate these issues, a subsequent step involves expanding the size of the SSM by incorporating more shapes to better capture higher inter-individual variability in anatomy. Furthermore, the mean shape should undergo expert annotation to establish a foundational reference, thereby ensuring the accuracy of the annotation on the generated training data. Ideally, the methodology presented in this paper could be applied to identify the same anatomical landmarks on 3D reconstructed laparoscopic images. This ensures consistent annotation across both modalities, facilitating 3D-3D registration using anatomical landmarks.
\\
Concluding, our approach demonstrates the potential of leveraging a comprehensive labeled training dataset using an SSM for deep-learning based 3D anatomical landmarks segmentation.

\begin{credits}
\subsubsection{\ackname}This study was funded
by Olympus Surgical Technologies Europe.

\end{credits}
%
%
%

\begin{thebibliography}{8}
\bibitem{pointnet}
Qi, Charles R., et al. "Pointnet: Deep learning on point sets for 3d classification and segmentation." \textit{Proceedings of the IEEE conference on computer vision and pattern recognition.} 2017. \doi{https://doi.org/10.48550/arXiv.1612.00593}

\bibitem{pointnet+}
Qi, Charles Ruizhongtai, et al. "Pointnet++: Deep hierarchical feature learning on point sets in a metric space." \textit{Advances in neural information processing systems} 30 (2017).

\bibitem{runnan}
Chen, Runnan, et al. "Semi-supervised anatomical landmark detection via shape-regulated self-training." \textit{Neurocomputing} 471 (2022): 335-345.

\bibitem{labrunie}
Labrunie, M., et al. "Automatic preoperative 3d model registration in laparoscopic liver resection." \textit{International Journal of Computer Assisted Radiology and Surgery} 17.8 (2022): 1429-1436. \doi{10.1007/s11548-022-02641-z}

\bibitem{koo}
Koo, Bongjin, et al. "Automatic, global registration in laparoscopic liver surgery." \textit{International Journal of Computer Assisted Radiology and Surgery} (2022): 1-10.

\bibitem{kokko}
Kokko, Michael A., et al. "Modeling the surgical exposure of anatomy in robot-assisted laparoscopic partial nephrectomy." \textit{Medical Imaging 2020: Image-Guided Procedures, Robotic Interventions, and Modeling.} Vol. 11315. SPIE, 2020.

\bibitem{tom}
François, Tom, et al. "Detecting the occluding contours of the uterus to automatise augmented laparoscopy: score, loss, dataset, evaluation and user study." \textit{International journal of computer assisted radiology and surgery} 15 (2020): 1177-1186.

\bibitem{ambellan19}
Ambellan, Felix, et al. \textit{Statistical shape models: understanding and mastering variation in anatomy.} Springer International Publishing, 2019.

\bibitem{schaufel}
Schaufelberger, Matthias, et al. "A Radiation-Free Classification Pipeline for Craniosynostosis Using Statistical Shape Modeling." \textit{Diagnostics} 12.7 (2022): 1516.

\bibitem{nagel}
Nagel, Claudia, et al. "A bi-atrial statistical shape model for large-scale in silico studies of human atria: model development and application to ECG simulations." \textit{Medical Image Analysis} 74 (2021): 102210.

\bibitem{atkins}
Atkins, Penny R., et al. "Prediction of femoral head coverage from articulated statistical shape models of patients with developmental dysplasia of the hip." \textit{Journal of Orthopaedic Research} 40.9 (2022): 2113-2126.

\bibitem{mofrad}
Mofrad, Farshid Babapour, and Gelareh Valizadeh. "DenseNet-based transfer learning for LV shape Classification: Introducing a novel information fusion and data augmentation using statistical Shape/Color modeling." \textit{Expert Systems with Applications} 213 (2023): 119261.

\bibitem{Elharrouss}
Elharrouss, Omar, et al. "3D Point Cloud for Objects and Scenes Classification, Recognition, Segmentation, and Reconstruction: A Review." \textit{Cloud Computing and Data Science} (2023): 134-160.

\bibitem{krnjaca}
Krnjaca, Denis, et al. "A Statistical Shape Model Pipeline to Enable the Creation of Synthetic 3D Liver Data." \textit{Current Directions in Biomedical Engineering.} Vol. 9. No. 1. De Gruyter, 2023. \doi{10.1515/cdbme-2023-1035}

\bibitem{sprin}
You, Yang, et al. "Prin/sprin: On extracting point-wise rotation invariant features." \textit{IEEE Transactions on Pattern Analysis and Machine Intelligence} 44.12 (2021): 9489-9502. \doi{10.1109/TPAMI.2021.3130590}

\bibitem{ircad}
Soler, Luc, et al. "3D image reconstruction for comparison of algorithm database." URL: \url{https://www.ircad.fr/research/datasets/liver-segmentation-3d-ircadb-01 3} (2010).

\bibitem{adam}
Kinga, D., and Jimmy Ba Adam. "A method for stochastic optimization." \textit{International conference on learning representations (ICLR).} Vol. 5. 2015.

\bibitem{medshapenet}
Li, Jianning, et al. "MedShapeNet--A Large-Scale Dataset of 3D Medical Shapes for Computer Vision." \textit{arXiv preprint arXiv:2308.16139} (2023).

\bibitem{pytorch}
Paszke, Adam, et al. "PyTorch: An imperative style, high-performance deep learning library." \textit{Advances in neural information processing systems, 32}. (2019)

\end{thebibliography}
%

\end{document}